\documentclass[letterpaper, 10pt, conference]{ieeeconf}  %

\IEEEoverridecommandlockouts                              %

\usepackage{algorithmic, adjustbox, amsmath, amssymb, caption, cite, color, floatrow, graphicx, multirow, setspace, subcaption, xcolor, xspace}
\newfloatcommand{figurebox}{figure}[\nocapbeside][\dimexpr(\textwidth-\columnsep)/2\relax]
\newfloatcommand{tablebox}{table}[\nocapbeside][\dimexpr(\textwidth-\columnsep)/2\relax]

\usepackage[linesnumbered,algoruled,boxed,lined]{algorithm2e}
\SetKwComment{Comment}{$\triangleright$\ }{}

\SetCommentSty{xCommentSty}
\SetSideCommentRight
\DontPrintSemicolon

\title{\LARGE \bf
Monocular Plan View Networks for Autonomous Driving%
}

\author{Dequan Wang$^{1}$, Coline Devin$^{1}$, Qi-Zhi Cai$^{2}$*, Philipp Kr\"ahenb\"uhl$^{3}$, Trevor Darrell$^{1}$%
\thanks{$^{1}$UC Berkeley, $^{2}$Nanjing University, $^{3}$UT Austin}%
\thanks{*Work done while at UC Berkeley}%
}

\begin{document}

\newcommand{\ra}[1]{\renewcommand{\arraystretch}{#1}}
\newcommand{\ver}[1]{\rotatebox{90}{#1}}

\newcommand{\pk}[1]{\textcolor{brown}{PHILIPP: #1 }}
\newcommand{\dq}[1]{\textcolor{orange}{DEQUAN: #1 }}
\newcommand{\cd}[1]{\textcolor{cyan}{[COLINE: #1 ]}}
\newcommand{\td}[1]{\textcolor{magenta}{TREVOR: #1 }}
\newcommand{\todo}[1]{\textcolor{red}{[TODO: #1 ]}}

\newcommand{\etal}{\textit{et al}. }
\newcommand{\ie}{\textit{i}.\textit{e}., }
\newcommand{\eg}{\textit{e}.\textit{g}. }

\maketitle

\begin{abstract}

Convolutions on monocular dash cam videos capture spatial invariances in the image plane but do not explicitly reason about distances and depth.
We propose a simple transformation of observations into a bird's eye view, also known as plan view, for end-to-end control.
We detect vehicles and pedestrians in the first person view and project them into an overhead plan view.
This representation provides an abstraction of the environment from which a deep network can easily deduce the positions and directions of entities.
Additionally, the plan view enables us to leverage advances in 3D object detection in conjunction with deep policy learning.
We evaluate our monocular plan view network on the photo-realistic Grand Theft Auto V simulator.
A network using both a plan view and front view causes less than half as many collisions as previous detection-based methods and an order of magnitude fewer collisions than pure pixel-based policies.
\end{abstract}

\section{Introduction}

Autonomous driving promises accessibility, convenience, and safety on the road. 
Some recent approaches for training driving policies use demonstration data from monocular RGB images~\cite{bojarski2016end,xu2017end,codevilla2017end}. 
However, the large observation space of monocular RGB images may introduce many correlations that do not hold over in an on-policy evaluation~\cite{codevilla2018offline}. 
Moreover, a convolutional network assumes the wrong invariances when used on a first-person image. For example, a ConvNet is invariant to image plane translation but not to perspective or rotation. While these invariances are useful for classification tasks, relative and absolute positions matter in driving.
A tree on the side of the road can be safely ignored, 
while a tree in the path of the vehicle quickly leads to an accident.
Perspective effects in an image ensure that a plain monocular ConvNet needs to learn different filters for each object in each position and distance from the car.

The autonomous driving task requires reasoning about free space and other drivers' future positions. 
Classical navigation methods tend to use range-finders to draw a map of free space, 
 then compute controls from the top-down map. 
As shown by Tamar \etal, ConvNets can perform value iteration and planning on top-down navigation environments~\cite{tamar2016value}. 
To avoid relying entirely on LiDAR sensors, 
which are expensive and may misread rain or transparent surfaces, 
many modern approaches to deep learning for autonomous driving use first-person RGB views of the road. These are naturally available from cameras on a car but which are a less well structured representation of the state. 

While a first-person view image provides information about object positions and depths, the free space and overall structure of the scene is implicit rather than explicit. 
Hallucinating a top-down view of the road makes it easier to learn to drive
as free and occupied spaces are explicitly represented at a constant resolution through the image. 
A top-down view also can include the agent into the observation.

\begin{figure}[t]
    \begin{center}
        \includegraphics[width=\linewidth]{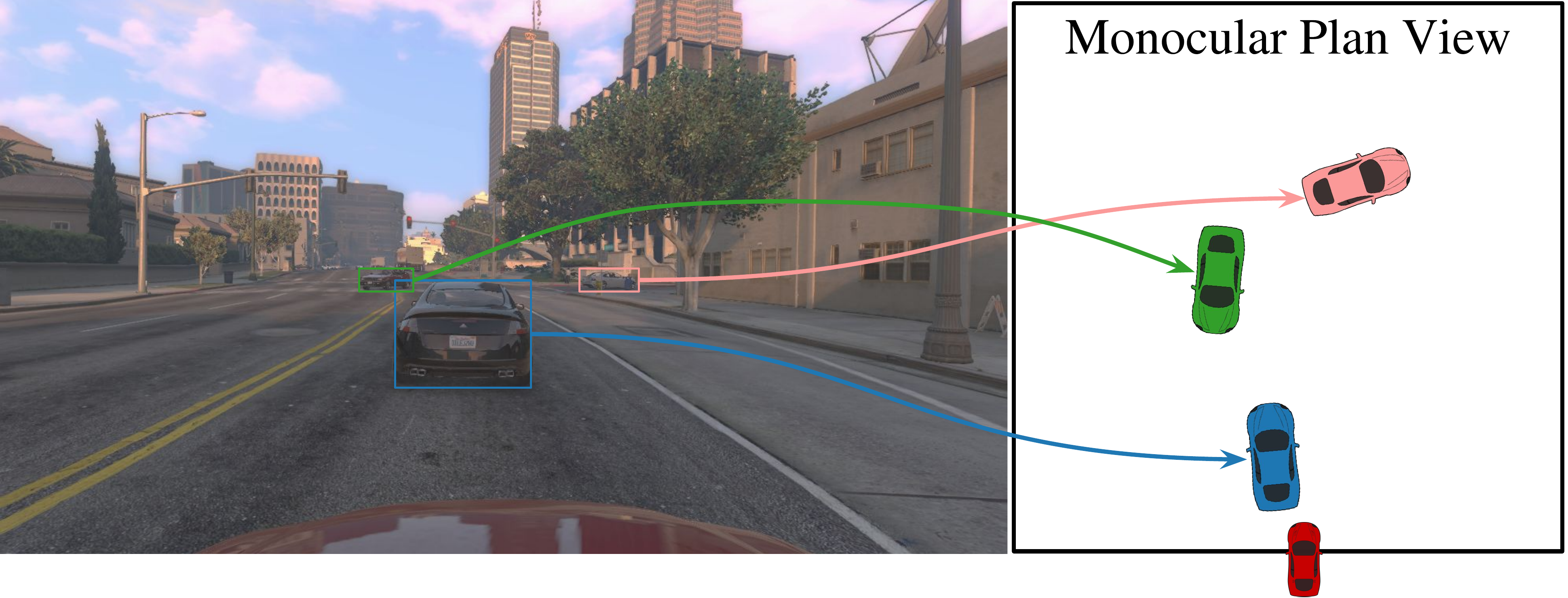}
    \end{center}
    \vspace{-2mm}
    \caption{Monocular Plan View Networks (MPV-Nets) use a monocular plan view image together with a first-person image to learn a deep driving policy.
    The plan view image is generated from the first-person image using 3D detection and reprojection.}
    \label{fig:concept}
\end{figure}

\begin{figure*}[t]
    \begin{center}
        \includegraphics[width=1\linewidth]{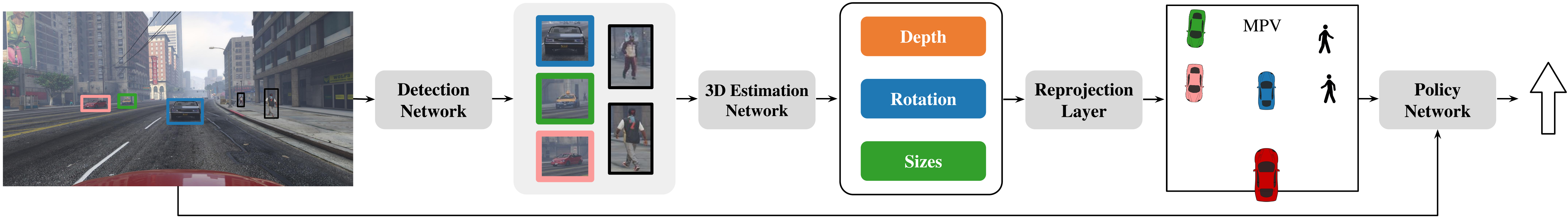}
    \end{center}
    \vspace{-2mm}
    \caption{Overview of Monocular Plan View Networks (MPV-Nets). The 2D detector and 3D estimation modules are trained with intermediate supervised losses. The Monocular Plan View (MPV) image is generated with geometric reprojection using the output of the 3D estimation. The final driving policy combines convolutional networks on the  first-person view and MPV to predict a driving action.}
    \label{fig:overview}
\end{figure*}

Intuitively, two- and three-dimensional object detectors~\cite{mousavian20173d} should provide useful information for autonomous driving. However, despite the advances in 3D object detection~\cite{chen2016monocular,mottaghi2015coarse,xiang2017subcategory}, few attempts have been made to use these detections in conjunction with deep learning in robotics.
Prior methods included them either in the monocular view or with an attention based pooling~\cite{wang2018deep,devin2017deep}.
Both approaches struggle to reason about the intricate three dimensional interplay between cars in complex traffic patterns.
Our approach leverages the spatial reasoning power of ConvNets to analyze and react to these traffic patterns.

In this paper, we propose Monocular Plan View Networks (MPV-Nets) to combine the advantages of first-person views, which are naturally available from the driving agent, and top down views, which enable reasoning about objects and space. The MPV-Nets require only a monocular first-person image and infers the depths and poses of the objects within the scene. With this 3D information, the MPV-Nets reproject the objects into an overhead plan view.

This top-down image allows the MPV-Nets policy to easily extract and use features related to spatial positions of objects in the scene. 
A two dimensional ConvNet learns to predict an action from this overhead view.
This low-dimensional representation emphasizes free-space around the agent 
in contrast to first-person images, where weather or style variations may confuse the policy.

Our entire pipeline is end-to-end trained to imitate expert trajectories with intermediate detection losses.
We use detection and 3D estimation supervision for each object during training and infer these during testing.
We evaluate our driving policy on the photo-realistic Grand Theft Auto V simulator.
Our experiments show that an MPV policy performs significantly better than standard monocular policies with and without objects detections.
The MPV policy both travels farther and experiences fewer collisions and interventions than front-view policies. Our model also outperforms using only the plan view, indicating that the front view image pixels do provide necessary information.

\section{Related Work}

\begin{figure*}[t]
    \begin{center}
        \includegraphics[width=1\linewidth]{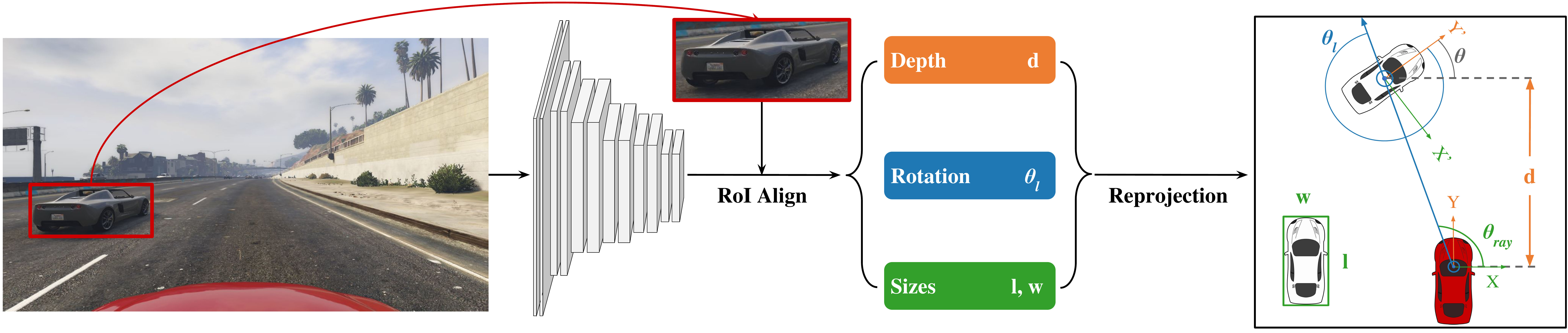}
    \end{center}
    \vspace{-2mm}
    \caption{The 3D estimation module uses ROIAlign features from the image to infer the plan view location, orientation and dimensions of each object. This information is used to render objects into the plan view.}
    \label{fig:plan_view}
\end{figure*}

Our inferred monocular plan view representation has some similarities with occupancy grids generated by LiDAR sensors: 
we contrast free space with space occupied by vehicles and pedestrians. 
Recent works have used convolutional networks with LiDAR data for localization~\cite{barsan2018learning}, scene flow estimation~\cite{ushani2018feature}, 3D detection and tracking~\cite{chen2017multi, luo2018fast, yang2018pixor, frossard2018end, dequaire2018deep, yang2018hdnet, liang2018deep}, intention prediction~\cite{casas2018intentnet}, and semantic segmentation~\cite{zhang2018efficient}.
In particular,
Luo \etal propose a joint framework for 3D detection, tracking, and motion forecasting via 3D convolution over a bird's eye view representation of 3D world~\cite{luo2018fast}.
Yang \etal propose a map-aware single-stage 3D detector on the LiDAR bird's eye view~\cite{yang2018hdnet}.
Petrovskaya \etal use the concept of a ``virtual scan", which is a 2D representation of the 3D rangefinder output and which accounts for the ego-motion of the car~\cite{petrovskaya2008model}. 
Darrell \etal construct a plan view model from multiple cameras to track objects from a stereoscopic 3D estimate in order to help track entities and obtain a globally consistent reference frame~\cite{darrell2001plan}.

Our work, on the other hand, uses the plan view as new input modality to a deep network.
Our work differs as we only use an RGB camera sensor and specifically detect vehicles and pedestrians in order to construct the view.
Palazzi~\etal generate top-down vehicle coordinates from a first-person RGB input but require paired top and first-person images and do not use these in a driving policy~\cite{palazzi2017learning}. 
Our method only re-projects detected objects and can be trained from monocular video and three dimensional location or from tracking information.

Training neural network policies for driving has a long history.
ALVINN~\cite{pomerleau1989alvinn} was the first usage of neural network for end-to-end driving from image data.
Muller \etal introduce a ConvNet-based obstacle avoidance system for off-road mobile robots~\cite{muller2006off}. 
More recently, Bojarski \etal\cite{bojarski2016end, bojarski2017explaining} and Xu \etal\cite{xu2017end} revisited end-to-end imitation learning, training a policy to output steering commands directly form RGB images.
Muller \etal added a bias to the deep policies by training a policy on the semantic segmentation of the scene as well as the image~\cite{muller2018driving}. 
The segmentation module enabled better transfer from synthetic to real, but does not include information about the 3D layout of the scene. 

Another form of domain knowledge was used by
Chen \etal\cite{chen2015deepdriving} and
Sauer \etal\cite{sauer2018conditional}, who chose a number of ``affordances" to predict and supervise, such as the distance between cars and lanes, or the status of the streetlights.
It is clear that adding some domain knowledge or model bias can significantly improve driving performance and transfer. However, supervising specific affordances can be an ad-hoc process. Our approach instead provides 3D information about the scene in a representation that showcases spatial relations between objects.

\section{Monocular Plan View Networks}

Figure~\ref{fig:overview} shows an overview of our proposed MPV-Net.
Given a monocular first person view RGB image, a 2D detector finds the image coordinates of all objects in the scene.
The 3D estimation network then computes the depth, rotation, and 3D size of each object.
A reprojection layer renders each 3D object into a plan view image using the 3D bounding box information. 
The original first-person image and the generated plan view are each passed through separate convolutional networks and globally pooled into two feature vectors. 
The resulting features are concatenated and fed to a fully connected layer which outputs a discrete action. This pipeline is summarized in Algorithm~\ref{alg:mpv}.

\paragraph{2D detection}
Our goal is to construct and utilize a plan view for policy learning.
However, our observation space is a simple first-person view image $I_{FV}$ from a dash camera of a car.
In order to construct the plan view we first detect all cars and pedestrians in an image.
We use an off-the-shelf Mask RCNN~\cite{he2017mask} detector to extract a 2D bounding box $(u,v)_k$ for each object $k$.
Next, we estimate the 3D position, size, and heading of each object.

\paragraph{3D estimation}

The 3D module estimates the object's 3D pose: plan view 2D position $(X, Y)$, angle in plane (yaw) $\theta$, and size $(l,w,h)$.
As input, the 3D module takes a bounding box $(u,v)_i$ and the front view image $I_{FV}$.
Inspired by Mousavian~\etal\cite{mousavian20173d}, we train a network $E_{3D}$ with three sub-networks to estimate the depth $d$, local orientation $\theta_l$, and size $(l, w, h)$ of each object.
Figure~\ref{fig:plan_view} shows an overview of the module.
All three sub-networks share a common fully convolutional base network.
The base network extracts dense features from the original input image.
For each bounding box, we extract a 3D feature $f_k$ from the base network using ROIAlign~\cite{he2017mask}.
The three sub-networks then predict their corresponding outputs from these ROIAlign features:
\begin{align}
(d, \theta_l, l, w, h)_k = E_{3D}(f_k)\text{ for object } k
\end{align}
In addition, we compute the relative horizontal position $x_k$ from the center of the 2D bounding box.

For a particular object, we compute the global rotation $\theta$ in the camera coordinate system from $\theta_l$, as shown in Figure~\ref{fig:plan_view}.
Given the horizontal distance from the vertical center of image $x$ and the camera's focal length $f$, 
\begin{align}
    \theta = (\theta_l - \arctan \frac{x}{f})\mod 2\pi.
\end{align}
The relative plan view location of the object is given by
\begin{align}
    X = \frac{d x}{f}, \quad\quad Y = d.
\end{align}
The width $w$ and length $l$ of an object are directly estimated in plan-view space. We obtain the ground truth focal length $f$ directly from the simulator and learn all network parameters with direct supervision.

Our 3D estimation network $E_{3D}$ does not share a base network with our detector for two reasons:
First, it significantly simplifies training.
Second, the two networks learn drastically different features.
For example, a detection network should be scale invariant, which works well for size and rotation estimates, but provides little information about the depth of an object.
However, in order to perform good monocular depth estimation, the base network needs to detect the absolute 2D size of objects and store it in an ROI feature.
Only a prior on the size of objects will yield a good depth estimate.
A good depth estimation network also heavily exploits priors over object locations, which a detector should ignore.

\begin{algorithm}[htp]
\SetAlgoLined
\footnotesize
\KwIn{front-view image $I$}
   Initialize plan view $I_{PV}$ to zeros\;
   $(u, v)_1, ..., (u, v)_k \gets \text{D}(I_{FV})$ \tcc*{2D detection}
  \For{$i \in [1, k]$}{
   $X, Y, \theta, l, w, h \gets E_{3D}((u, v)_i, I_{FV})$ \\
   \tcc*{3D estimation}
   $o_{PV} \gets
\begin{bmatrix}
X\\
Y\\
\end{bmatrix}
+
R(\theta) \begin{bmatrix}
\pm \frac{w}{2}\\
\pm \frac{l}{2}\\
\end{bmatrix}$ \tcc*{Reprojection}
   $I_{PV}[o_{PV}] = 1$ \tcc*{Rendering}
  }
  $f_{FV} \gets C_{FV}(I_{FV})$ \tcc*{Front view features}
  $f_{PV} \gets C_{PV}(I_{PV})$ \tcc*{Plan view features}
  $a = \pi(f_{FV}, f_{PV})$ \tcc*{Policy network}
  \KwOut{action $a$}
  \caption{An MPV policy.}
  \label{alg:mpv}
\end{algorithm}

\paragraph{Reprojection}
The reprojection layer converts the 3D estimates ($\theta$, $X$, $Y$, $l$, $w$) of all objects into a plan view image.
We create a separate plan view image for each class of objects.
The plan view image is assigned a value $1$ if any object occupies the spatial location corresponding to a pixel and $0$ otherwise.
For a particular object centered at $X$, $Y$ with size $l, w, h$ and rotation $\theta$, the reprojection layer computes the extent of the object in the plan view.
We describe each object in plan view as a rotated bounding box with four corners defined as
\begin{equation}
o_{PV} = 
\begin{bmatrix}
X\\
Y\\
\end{bmatrix}
+
R(\theta) \begin{bmatrix}
\pm \frac{w}{2}\\
\pm \frac{l}{2}\\
\end{bmatrix}
\end{equation}
where $R(\theta)$ is the rotation matrix for angle $\theta$.

There are several options to choose from when deciding how to structure the reprojection, 
such as what information to include in the view and what orientation to face. 
In this work, we focus on reprojecting vehicles and pedestrians, but one could also overlay a map if available, or detect street signs, lights, lanes, etc.
Here we explore two possible orientations of the plan view:
First, the scene is rendered relative to the agent with the agent always facing upwards as in Figure~\ref{fig:concept}. 
We call this type of MPV the direction-of-travel view.
This may lead to better generalization because driving south would look the same as driving north. 
However, this view is easily distracted by the agent's own actions and is not invariant to ego-motion and rotation.
The second option we explore uses a constant frame of reference by always placing north at the top. 
We name this north-up view.

Once we fix the reference frame of the plan view, we render each moving object (vehicle or pedestrian) into a top-down image.
We use an aliased rendering technique and set all pixels inside an object box to $1$.
This step is not differentiable but could easily be replaced with a differentiable anti-aliased alternative~\cite{loper2014opendr} if an end-to-end gradient is required.
Figure~\ref{fig:plan_view} shows the complete 3D module.

\begin{figure}[t]
    \begin{center}
        \includegraphics[width=1\linewidth]{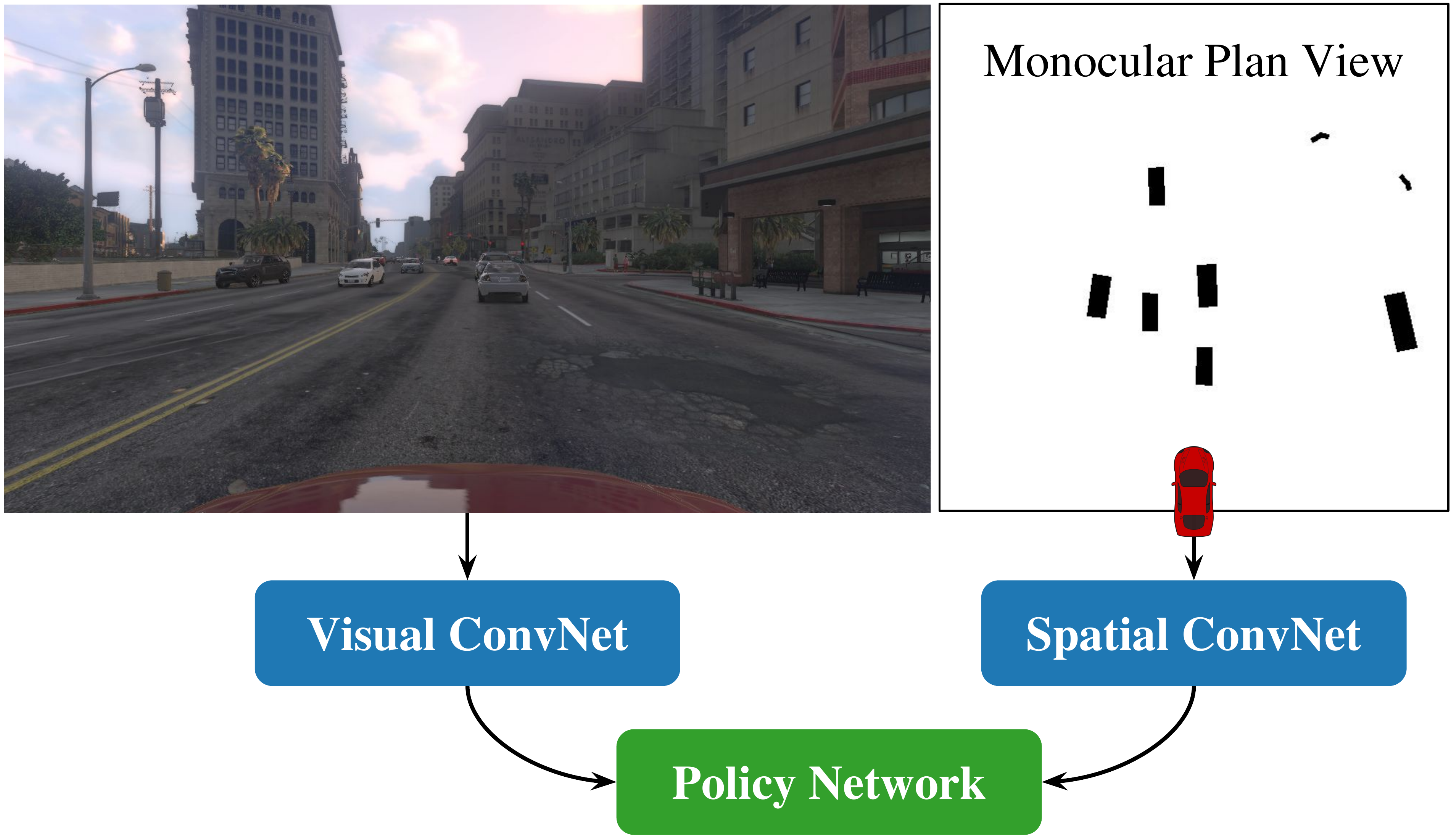}
    \end{center}
    \vspace{-2mm}
    \caption{Overview of our policy network. Our architecture uses both the first person view and our plan view to predict driving actions. Each image is passed through a convolutional network and pooled into a feature vector. The plan view and image vectors are concatenated, and passed through a fully connected layer to produce a discrete action.}
    \label{fig:policy_network}
\end{figure}

\paragraph{Policy Network}

Given the generated MPV image,
we use a convolutional network to extract visual and spatial features for both RGB image and MPV images.
The visual ConvNet $C_{FV}$ is pre-trained on ImageNet~\cite{russakovsky2015imagenet}, while the spatial ConvNet $C_{PV}$ is trained from scratch.
The concatenation of both visual and spatial features is fed into a linear policy $\pi$ to produce actions. 
Figure~\ref{fig:policy_network} shows the policy network structure.
We train the the ConvNets and $\pi$ jointly with imitation learning.

\section{Experiments}
We evaluate MPV-Nets in a realistic simulated driving environment. All models are trained with imitation learning, and their driving performance is evaluated through several metrics, including the number of collisions and the distance traveled. The aim of the experiments is to determine whether explicit use of plan view does help learn driving policies as well  which representation of the 3D object is best. As the 3D detections are inferred from the same front view image used by the baseline models, no additional information is generated at test time by the detections. However, as seen in Figure~\ref{fig:onpolicy}, we find that this structured representation does improve final driving performance. We also show in Table~\ref{table:perplexity} that the 3D detections alone are not sufficient for driving: a policy trained solely on the MPV without the front view image features is unable to predict the expert's actions.

\subsection{Experimental Settings}

\paragraph{Dataset}
All models (baselines, related works, and our MPV-Nets) are trained on a dataset of (image, action) pairs generated from Grand Theft Auto V.
Leveraging the in-game navigation system as the expert policy, 
we collect 2.5 million training frames over $1000$ random paths over $2$km at 12 fps.
With the default cloudy weather, we collect driving demonstrations from 8:00 am to 7:00 pm.
The horizontal field of view (FoV) for the dash cam is fixed as fixed $60^\circ$.
Inspired by DAgger~\cite{ross2011reduction}, we randomly add noise to the expert's action every 30s to augment the data with examples of error recovery. The noisy control actions along the following seven frames are removed when saving the data for training to avoid imitating the noisy behavior.

Each frame includes a front-view RGB image and control signals such as speed, angle, throttle, steering, and brake.
We discretize the target control into 9 actions: (\textit{left, straight, right}) $\times$ (\textit{fast, slow, stop}). 
The 3D estimation module is also trained on the simulated data: we extract 3D ground-truth pose labels for each object relative to the agent. These are not used during evaluation unless the experiment is labeled ``ground truth".

\paragraph{Training the models with imitation learning}
All models are trained on the dataset to predict the expert's action at each frame using a cross-entropy loss on the discretized action space. For the models using 3D detection (MPV and FVD), the 3D detection module is first trained to predict the relative position, depth, and size of cars and pedestrians in the image. Then, the 3D detection is frozen, and the convolutional networks that extract global features from the front view image and re-projected image, as well as the policy layer, are trained on the behavioral cloning objective.

\paragraph{Off-policy Evaluation}
As an initial evaluation, we can compare the perplexities of different models. The perplexity, which calculates the negative log likelihood of the expert data under the distribution predicted by the model (lower is better), measures how well the model fits the data. Lower perplexity does not always indicate high on-policy performance, but high perplexity (as in Table~\ref{table:perplexity}, bottom rows) indicates that the model has underfit. 

\begin{figure*}[ht]
	\centering
	\ra{1.2}
	\adjustbox{max width=0.95\columnwidth}{
		\begin{tabular}{c c c c c c}
		    \resizebox{0.5in}{!}{\ver{(a) Input Image}} &
			\includegraphics[width=\linewidth]{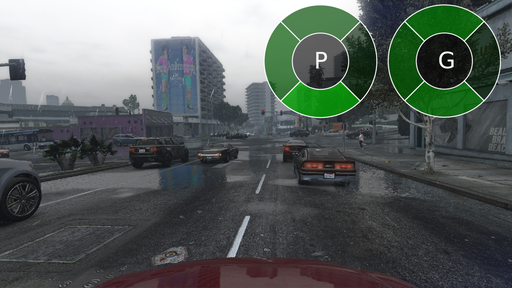} & 
			\includegraphics[width=\linewidth]{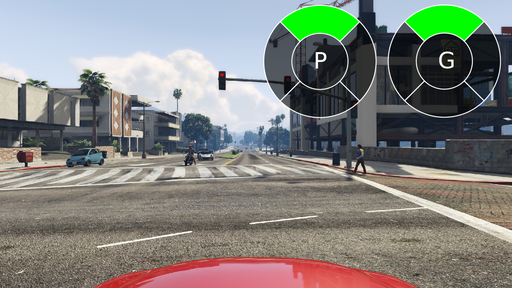} & 
			\includegraphics[width=\linewidth]{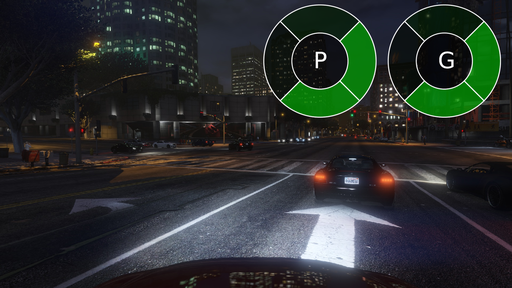} & 
			\includegraphics[width=\linewidth]{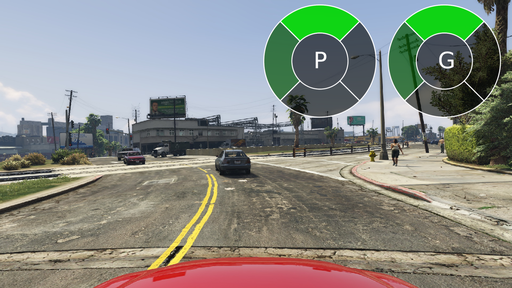} &
			\includegraphics[width=\linewidth]{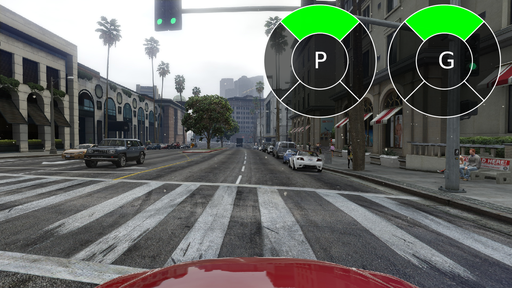} \\
            \resizebox{0.5in}{!}{\ver{(b) Prediction}} &
			\includegraphics[width=\linewidth]{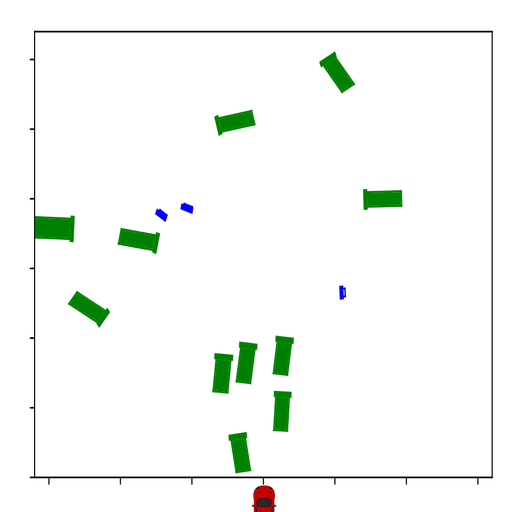} & 
			\includegraphics[width=\linewidth]{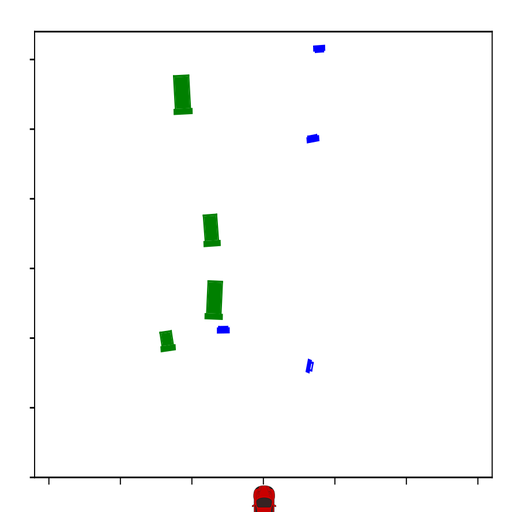} & 
			\includegraphics[width=\linewidth]{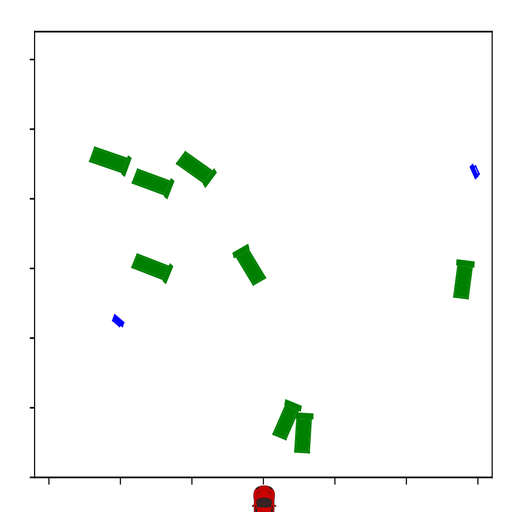} & 
			\includegraphics[width=\linewidth]{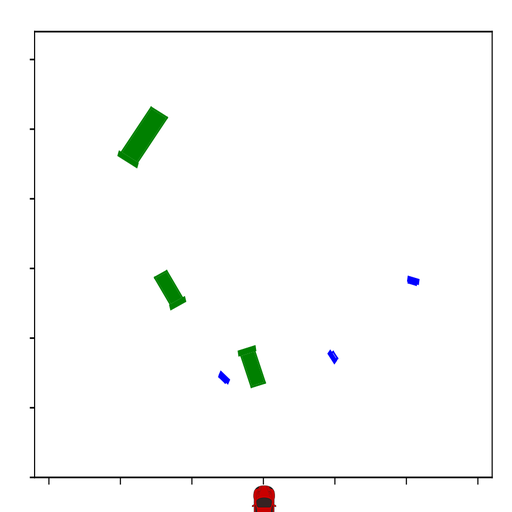} &
			\includegraphics[width=\linewidth]{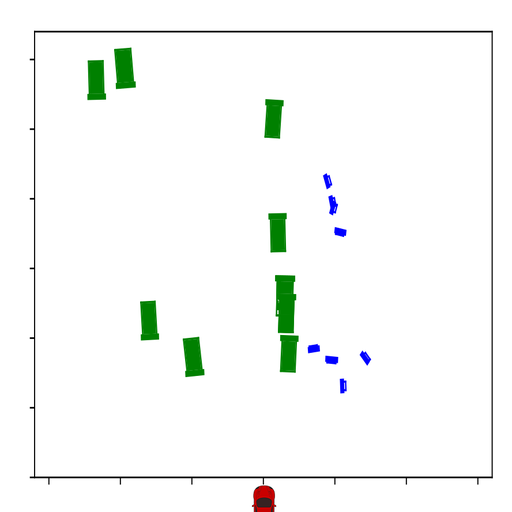} \\ 
            \resizebox{0.5in}{!}{\ver{(c) Ground Truth}} &
			\includegraphics[width=\linewidth]{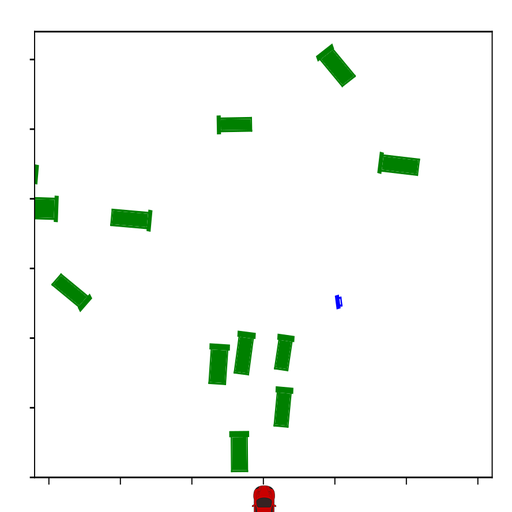} & 
			\includegraphics[width=\linewidth]{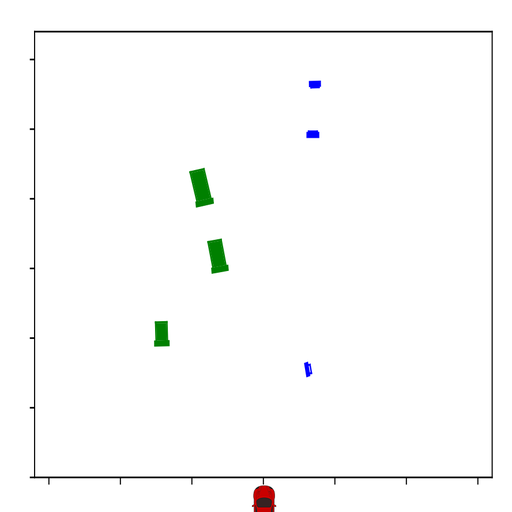} & 
			\includegraphics[width=\linewidth]{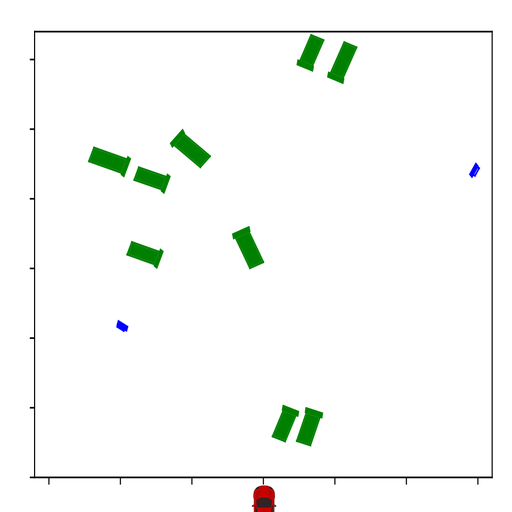} & 
			\includegraphics[width=\linewidth]{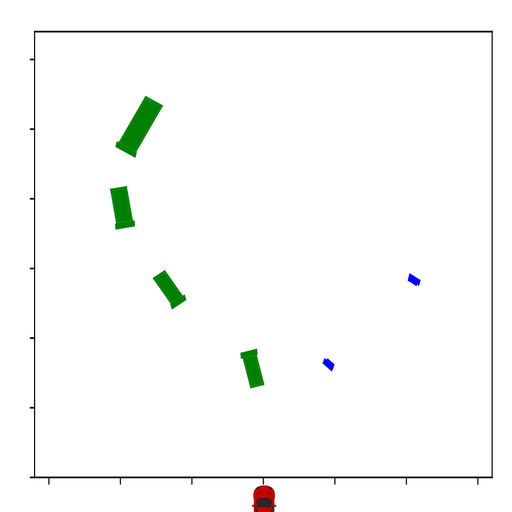} &
			\includegraphics[width=\linewidth]{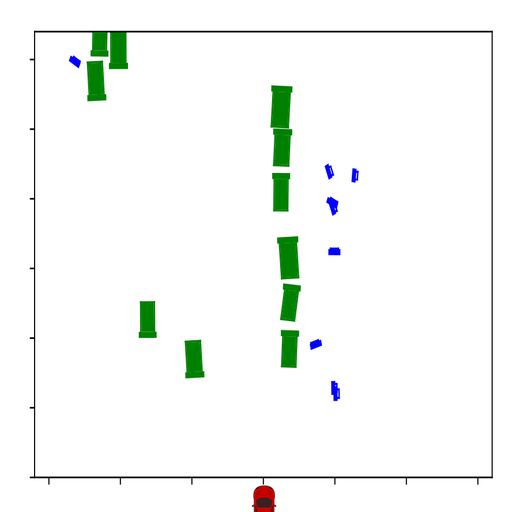} \\
		\end{tabular}}
		\caption{Sample scenes from the Grand Theft Auto V simulation along with our generated monocular plan view. Row (b) shows the rendering using the learned 3D detector, while (c) uses the ground truth object positions and headings. Vehicles are rendered in green and pedestrians in blue. The red car below each plot indicates the agent's relative position. 
		In the top row, the circles on the image indicate the action taken by the model using predicted 3D locations (P) and the one using ground-truth locations (G). Although the learned detector does not always detect every object, the detections are good enough for the driving policy: both models output the same actions for the randomly chosen scenes. Best viewed in color.
		}
		\label{fig:selection}
\end{figure*}

\paragraph{On-policy Evaluation}
We evaluate the models in $8$ locations unseen during training: $2$ highway and $6$ urban intersections. 
Figure~\ref{fig:selection} demonstrates some example scene layouts in our simulation environment.
For each test location, we run the model for 10 independent roll-outs lasting around $800$ steps, 10 minutes of game time, each. Because the model predicts discretized actions,  a hand-tuned PID controller (shared for all models) is used to convert the discrete actions into continuous control signals. Collisions are counted as detected by the environment. If the agent gets stuck or stops moving for $30$ seconds, we let the in-game AI intervene and count the intervention. Not all collisions lead to interventions. The time and distance traveled by the expert during an intervention are not counted in the evaluation.

The models are evaluated with several metrics. 
For each roll-out, we count the number of collisions, the number of interventions, and the distances traveled.
To aggregate across roll-outs, we compute the average distance driven between AI interventions, the number of collisions per 100m, and the number of interventions per 100m.

\begin{table*}[!t]
\ra{1.2}
\adjustbox{width=\linewidth}{
    \begin{tabular}{l|cccc|ccc|c|c}
    \hline
    \multirow{2}{*}{Object} & \multicolumn{4}{c|}{Depth Error Metric} & \multicolumn{3}{c|}{Depth Accuracy Metric} & \multicolumn{1}{c|}{Orientation Metric} & \multicolumn{1}{c}{Size Metric}\\ %
    & Abs Rel $\downarrow$ & Sq Rel $\downarrow$ & RMSE $\downarrow$ & RMSE$\log$ $\downarrow$ & $\delta < 1.25$ $\uparrow$ & $\delta < 1.25^2$ $\uparrow$ & $\delta < 1.25^3$ $\uparrow$ & OS $\uparrow$ & Dim $\uparrow$ \\ \hline
    Pedestrian & 0.059 & 0.381 & 4.014 & 0.094 & 0.970 & 0.994 & 0.998 & 0.873 & 0.968 \\
    Vehicle & 0.102 & 1.043 & 8.259 & 0.142 & 0.935 & 0.983 & 0.994 & 0.945 & 0.889 \\
    \hline
    \end{tabular}
}

\caption{Evaluation of our 3D estimation module on standard depth, orientation and size error metrics. An up arrow ($\uparrow$) indicates higher is better, a down arrow($\downarrow$) indicates lower is better. The accuracy of our system is $90+\%$ on most metrics.}
\label{tab:3d_estimation}
\end{table*}

\begin{figure}[t]
	\centering
	\ra{1.2}
	\includegraphics[width=\linewidth]{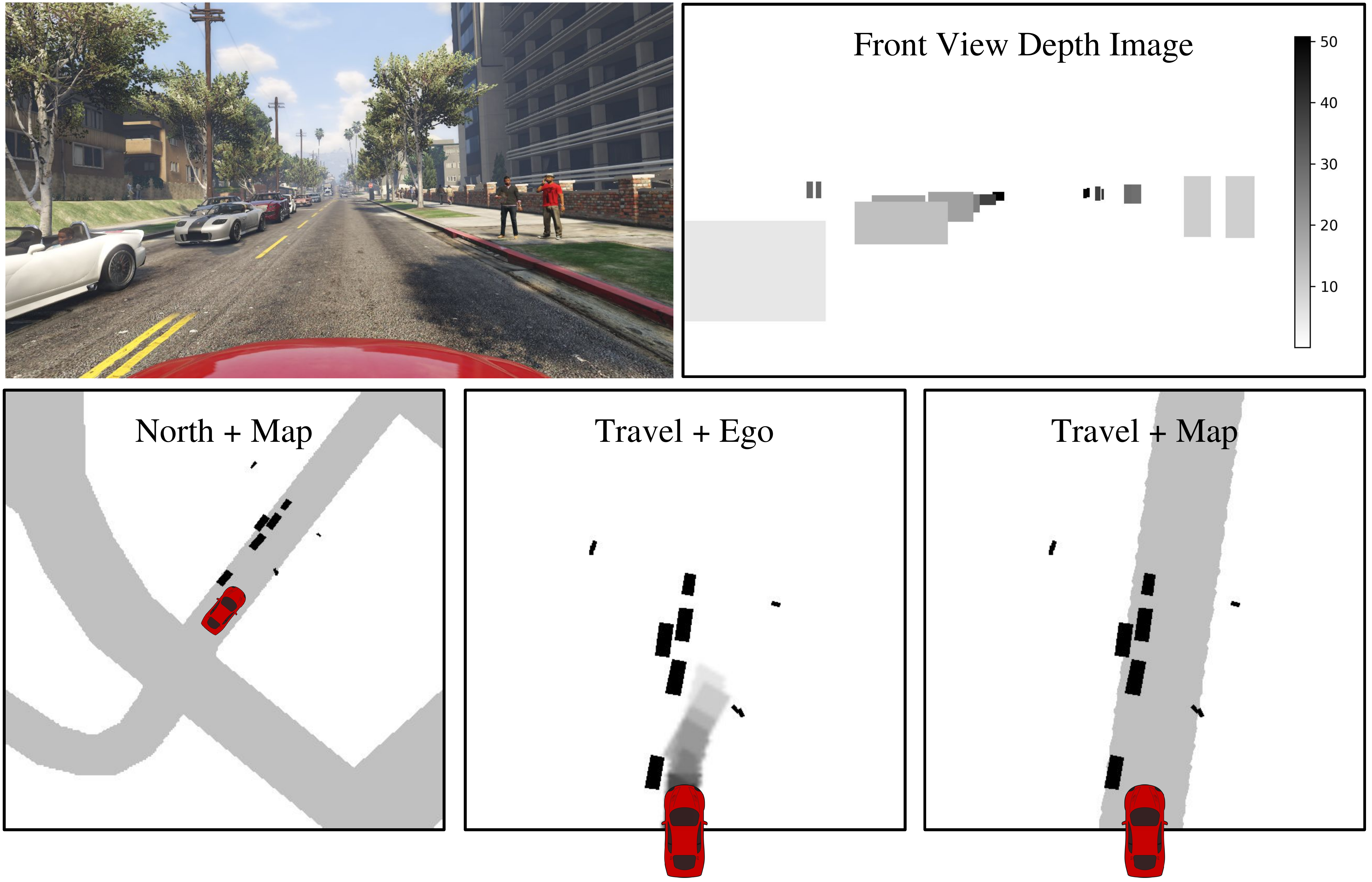}
	\caption{The top row shows a standard first person view (left) and an object based depth map (right) used for our \textit{depth} experiments. The bottom row shows design choices for the monocular plan view: a north-up view with map (left), direction of travel view and ego-history (center), direction of travel and map (right).  }
	\label{fig:design}
\end{figure}

\begin{table}[t]
	\ra{1.2}
	\adjustbox{max width=\linewidth}{
		\begin{tabular}{c c c c c c c c}
		    view & img & map & ego & perplexity & distance & interventions & collisions \\ 
            \hline
            travel & \checkmark & & & 0.989 & \textbf{268.95} & \textbf{1.30$\pm$0.78} & $3.38\pm2.55$\\
            north & \checkmark & & & 0.969 & 235.85 & $1.70\pm1.14$ & \textbf{2.46$\pm$2.18}\\
            travel & \checkmark & \checkmark & & 1.010 & 218.94 & $2.99\pm4.49$ & $4.12\pm4.42$\\
            north & \checkmark & \checkmark & & 0.975 & 213.79 & $1.92\pm3.37$ & $3.46\pm4.02$\\
            travel & \checkmark & & \checkmark & \textbf{0.834} & 144.92 & $2.94\pm1.79$ & $6.49\pm5.72$\\
            north & \checkmark & & \checkmark & 0.853 & 96.70 & $2.56\pm0.98$ & $6.60\pm6.32$ \\
            \hline
            travel & & & & 1.378 & - & - & - \\
            north & & & & 1.516 & - & - & - \\
            travel & & \checkmark & & 1.424 & - & - & - \\
            north & & \checkmark & & 1.516 & - & - & - \\
            
			\hline
		\end{tabular}}
		\caption{Ablation of different plan view models. We measure the perplexity (lower is better) of the models on withheld test data as an off-policy evaluation, as well as several on-policy performance metrics. Adding the agent's history or a map lead to a better off-policy evaluation, but have a poor driving performance. MPV without a map or history performs best. Travel view slightly outperforms the consistent north-up view. We also evaluate the policy network without the visual branch for image. The high test perplexities of these models indicate the underfit of MPV-only policy. Therefore our policy networks take both first-person view and top-down view.}
		\label{table:perplexity}
\end{table}

\subsection{Implementation}
We implement our system with the open-source deep learning framework PyTorch~\cite{paszke2017automatic}. 
We extend the work of Kr\"ahenb\"uhl~\cite{krahenbuhl2018free} and extract ground truth object 3D pose ground truth from Grand Theft Auto V. In order to provide a fair evaluation, all models predict 1 action every 7 frames, and the game is paused during the model's computation.
In this way, all the models are evaluated on the same number of forward passes.

For the 2D detection module,
we use a COCO~\cite{lin2014microsoft} pre-trained Mask RCNN~\cite{he2017mask} based on Detectron~\cite{Detectron2018}.
We use a 34-layer DLA model~\cite{yu2018deep} as the backbone architecture for 3D estimation and policy network.
The design of the 3D sub-network is the stack of three $3\times3$  convolutional layers.

We train our policy network with Adam \cite{kingma2014adam} for 2 epochs.
The initial learning rate is $0.001$ and the batch-size is $64$.
The original input RGB image is $1920\times1080$.
We resize the given image into $640\times352$ for both training and testing. 
The size of the MPV image is $512\times512$, spatially covering 64m$\times$64m of the environment.
We ignore the objects whose depth is greater than 64m for all experiments.
We only keep the objects within 32m to the right and left.
We do not use any visual data augmentation during training.

\subsection{Evaluation of 3D Esimation}
We evaluate the 3D estimation module for pedestrian and vehicle in Table~\ref{tab:3d_estimation}.
We utilize the depth and orientation evaluation metrics from Eigen \etal~\cite{eigen2014depth} and KITTI~\cite{geiger2012we}.
The depth estimation is evaluated by both error and accuracy metrics, 
while rotation is evaluated by orientation score (OS).
Size prediction is measured by the accuracy of the object volume estimation :
$Dim = \min(V_{\mathtt{pred}}/V_{\mathtt{gt}}, V_{\mathtt{gt}}V_{\mathtt{pred}})$
with an upper bound $1$, where $V = l w h$ is the volume of a 3D box.

Figure \ref{fig:selection} shows a comparison between predicted MPV images and the ground truth ones on randomly picked images.
The 3D module is robust to various scenarios under different weather and timing, but is not perfect. Common errors include missing detections or wrong orientations..
However, the imperfect perception system does not significantly impact the driving policy.
The policy predicts an almost identical distribution over actions for either predicted or ground truth MPV images.
The MPV policy network is robust to an imperfect perception module.

\newpage

\begin{figure*}[ht]
\begin{center}
    \includegraphics[width=\linewidth]{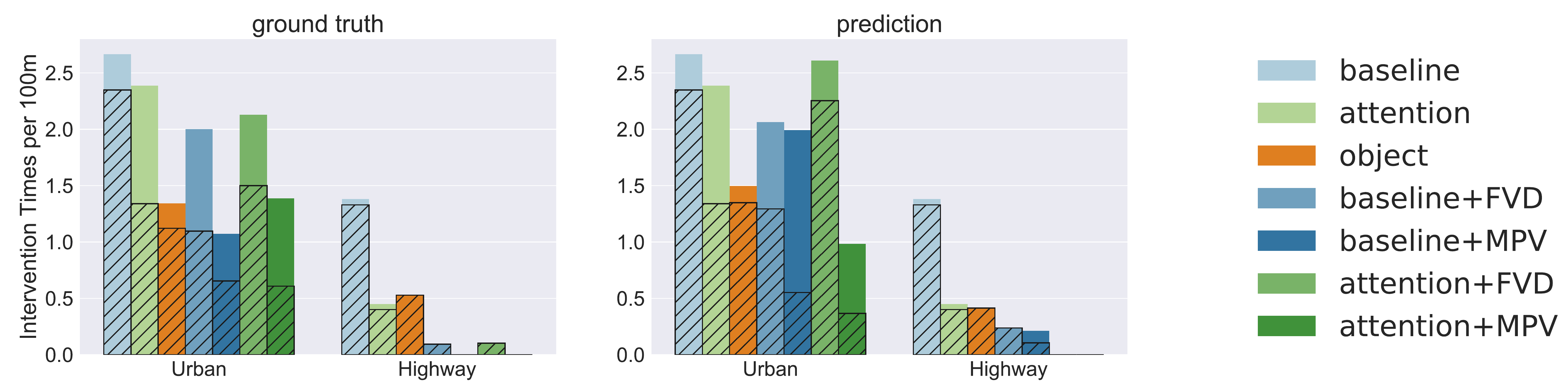} %
\end{center}
    \caption{Analysis of interventions for different policies. The shaded regions indicate the proportion of interventions that were caused by collisions instead of long stops. The \textit{attention+MPV} model both has the fewest interventions and fewest collisions.}
    \label{fig:interventions}
\end{figure*}
\begin{figure*}[t]
\centering
\includegraphics[width=\linewidth]{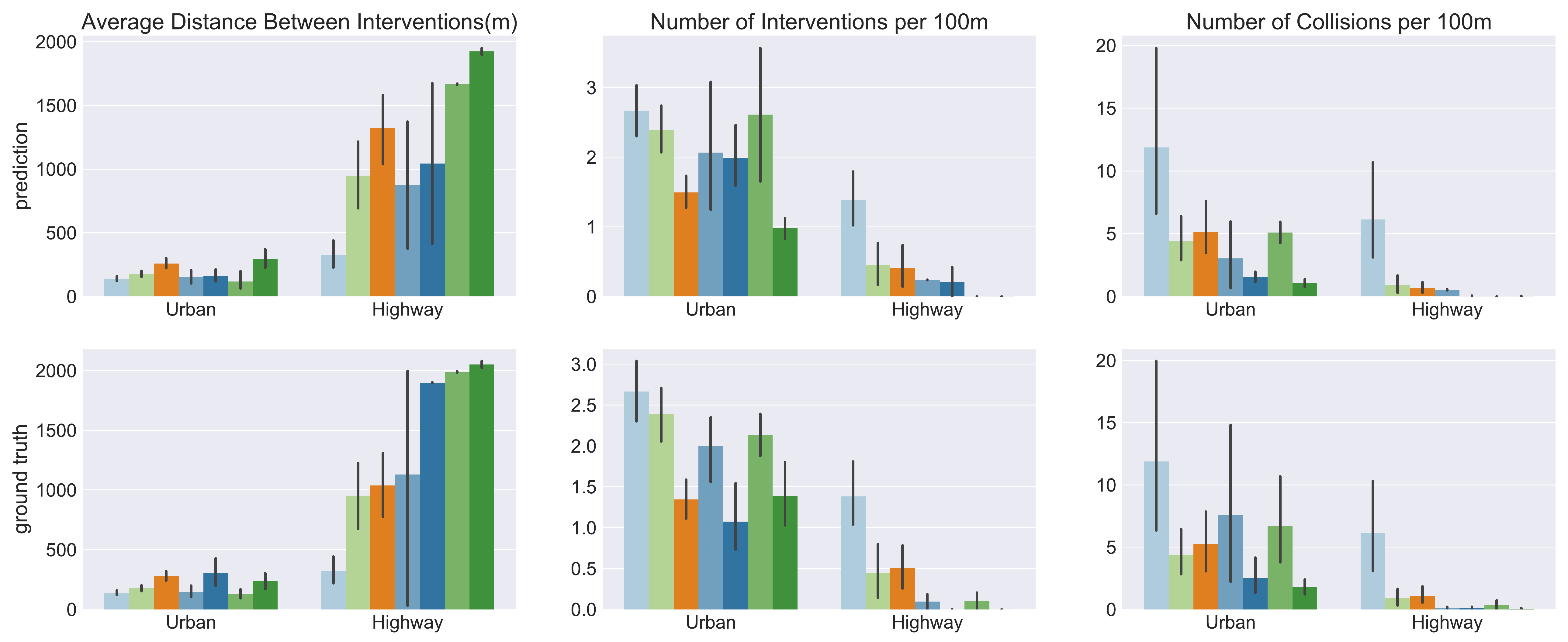}
\caption{Driving performance in urban and highway scenarios for different policies (see legend in Fig~\ref{fig:interventions}). From left to right: driving distance between interventions (higher is better), number of interventions per 100m (lower is better), number of collisions per 100m (lower is better). The models with MPV overall perform better than the baseline models (blue). MPV combined with attention performs best. }
\label{fig:onpolicy}
\end{figure*}
\begin{figure*}[ht]
    \begin{center}
        \begin{tabular}{c | c}
            \begin{subtable}[t]{0.46\textwidth}
            	\adjustbox{max width=\linewidth}{
                    \begin{tabular}{c c c c c c}
                        \includegraphics[width=\linewidth]{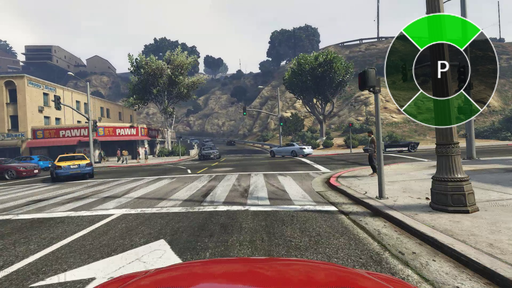} &
                        \includegraphics[width=\linewidth]{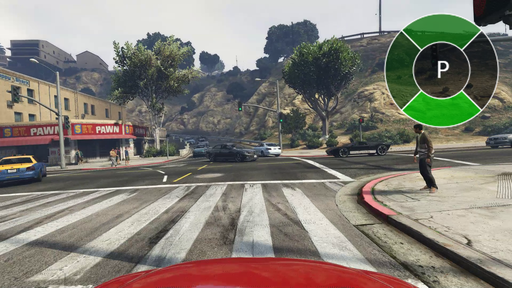} &
                        \includegraphics[width=\linewidth]{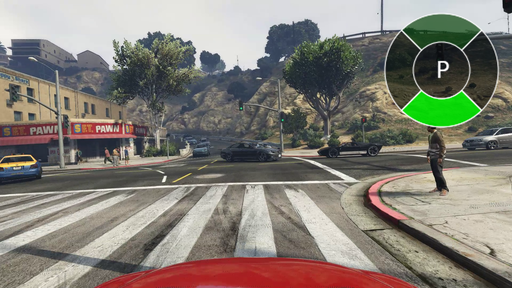} \\
                        \includegraphics[width=\linewidth]{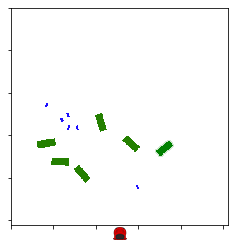} &
                        \includegraphics[width=\linewidth]{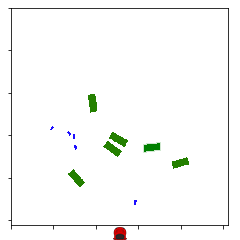} &
                        \includegraphics[width=\linewidth]{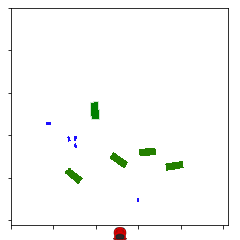} \\
            		\end{tabular}
            	}
                \caption{The agent stops as a pedestrian crosses the street.}
        	\end{subtable} 
        	&
        	
            \begin{subtable}[t]{0.46\textwidth}
            	\adjustbox{max width=\linewidth}{
                    \begin{tabular}{c c c c c c}
            			\includegraphics[width=\linewidth]{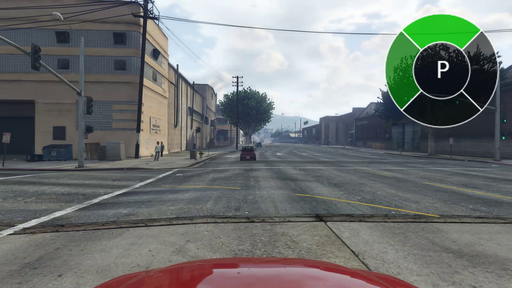} &
            			\includegraphics[width=\linewidth]{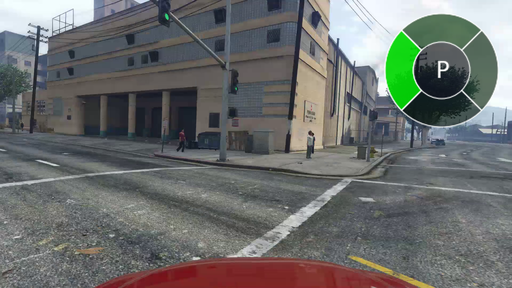} &
            			\includegraphics[width=\linewidth]{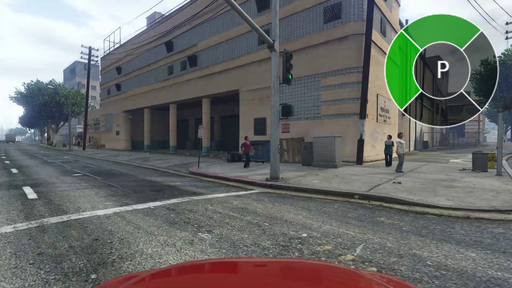} \\
                        \includegraphics[width=\linewidth]{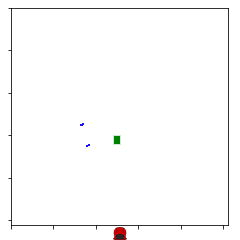} &
                        \includegraphics[width=\linewidth]{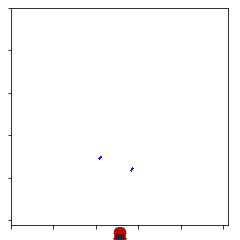} &
                        \includegraphics[width=\linewidth]{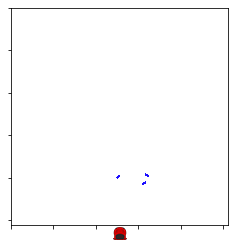} \\
            		\end{tabular}
            	}
            	\caption{The agent performs a U-Turn in a street with no other cars.}
        	\end{subtable} \\
        \end{tabular}
	\end{center}
	\caption{Visualization of action sequences of the attention+MPV policy using predicted 3D detections. The model is not trained for temporal consistency, and pose errors are independent for each frame. Trajectory (b) shows the importance of using both the plan view and the front-view RGB image: when very few objects are detected, the agent must rely the front view image for safe driving.}
	\label{fig:sequence}
\end{figure*}

\section{Results}
We evaluate all models on their on-policy driving performance.
For each model we compute the average distance between two interventions, the number of interventions per $100m$, and the number of collisions per $100m$.
The best models should drive long distances between interventions and have few interventions and collisions.

\paragraph{Design Choice Ablation}
We evaluate the design choices available for the plan view reprojection: (1) Rendering the plan view in direction of travel (Travel) of the agent or in a fixed north-up direction (North). (2) The plan view can include the ego position history of the agent within the image. (3) The plan view can include a coarse road map of the environment using the GPS position of the agent.
See Figure~\ref{fig:design} for a visual example.

Table~\ref{table:perplexity} shows the results in terms of off-policy perplexity, cross entropy of predicted actions on a held-out validation set, and our on-policy metrics.
The best models use neither a map nor the agent's ego motion. The orientation of the MPV is not important.
Including the agent's history performs much better in the off-policy perplexity evaluation but fails at on-policy driving.
This is due to the learning algorithm used: imitation learning, and specifically behavioral cloning.
It is prone to overfitting to the expert's data distribution, as supported by~\cite{muller2006off}. In contrast, the models trained on only the MPV without access to the front view image never learn to model the dataset. We were unable to test the on-policy performance of these models.

For all remaining experiments we use a direction-of-travel oriented plan view without a map or history. 

\paragraph{MPV-Net evaluation} 
We compare our model against several baselines, prior methods, and ablations. 
The \emph{baseline} method is based on the the network by Xu~\etal~\cite{xu2017end}, which uses a single convolutional network policy trained on the first-person view. 
The \emph{attention} method extends the \emph{baseline} with a pixel-level attention mechanism, learned end-to-end with the policy.
This is similar to Kim \etal~\cite{kim2017interpretable}.
Attention learns a weight mask for each feature location in the first-person view and MPV feature extractor.
This attention then guides a weighted average pooling.
The \emph{object} method builds on Wang \etal~\cite{wang2018deep}, and combines 2D object detections and deep image features, but lacks explicit depth supervision. 
We augment all our baselines with an additional front view depth (FVD) estimate without a plan view in the \textit{baseline+FVD} and \textit{attention+FVD} experiment.
See Figure~\ref{fig:design} for a visual example of the FVD.

Figure~\ref{fig:onpolicy} shows the main comparison on the on-policy evaluation.
A model that uses explicit depth information through either the MPV image or as an RGB-D image always improves performance when compared to the same model without depth.
Unsurprisingly, the simplest models (\textit{baseline} and \textit{attention}), which use no supervision other than the expert's actions, perform significantly worse than all other models.
This even holds for predicted depth estimates and detections.

We also find that, although the \textit{FVD} and \textit{MPV} are supervised with similar information, the \textit{MPV} models experience fewer interventions and collisions, especially in the busy urban environments.
This supports our hypothesis that MPV provides a better representation for policy learning.

To better understand each model's performance, we additionally plot the portion of interventions that are due to collisions in Figure~\ref{fig:interventions}.
We see that while the \text{attention+MPV} model has the fewest interventions, both MPV-based models have the lowest proportion of interventions caused by collisions in the urban setting.
This indicates that the plan view, more-so than simple depth estimation, is helpful for navigating through intersections.

Figure~\ref{fig:sequence} shows example rollouts of the attention+MPV-Net policy using the learned 3D estimator. In the top row, the plan view shows the pedestrian clearly in the way of the car, causing the car to brake appropriately. As we use the direction-of-travel reprojection, the detected entities rotate through the image when the car is turning, as shown in the second row of the figure.

\section{Discussion}
Monocular Plan View Networks (MPV-Nets) provide a top-down view of the driving environment using a standard front-view image and 3D object localization. 
Our results show that this representation improves driving performance in the GTA-V simulator, even when 3D detections are imperfect.
The quality of the driving policy does not degrade even when using noisy predicted 3D pose estimates.
By combining our MPV-Net with a pixel-level attention, we achieve almost perfect performance (0 collisions or interventions) on the highway driving task even when using predicted 3D detections. City intersections remain a challenge, but our plan view network provides promising results.
In comparison to traditional off-policy dataset evaluations, our on-policy evaluation provides a realistic understanding of how good a policy can be when trained from demonstrations.

MPV-Nets provide an intuitive way of incorporating trajectory history, maps, and object attention. While our ablations found that these additions hurt performance in the behavioral cloning setup, we expect that they would be very useful for planning or reinforcement learning based approaches. The plan view can intuitively and cleanly represent nearby vehicles' trajectories in a way that is accessible to a deep convolutional network. 

\begin{spacing}{1.008}
{%
\bibliographystyle{IEEEtran}
\bibliography{main.bib}

\begin{thebibliography}{10}
\providecommand{\url}[1]{#1}
\csname url@rmstyle\endcsname
\providecommand{\newblock}{\relax}
\providecommand{\bibinfo}[2]{#2}
\providecommand\BIBentrySTDinterwordspacing{\spaceskip=0pt\relax}
\providecommand\BIBentryALTinterwordstretchfactor{4}
\providecommand\BIBentryALTinterwordspacing{\spaceskip=\fontdimen2\font plus
\BIBentryALTinterwordstretchfactor\fontdimen3\font minus
  \fontdimen4\font\relax}
\providecommand\BIBforeignlanguage[2]{{%
\expandafter\ifx\csname l@#1\endcsname\relax
\typeout{** WARNING: IEEEtran.bst: No hyphenation pattern has been}%
\typeout{** loaded for the language `#1'. Using the pattern for}%
\typeout{** the default language instead.}%
\else
\language=\csname l@#1\endcsname
\fi
#2}}

\bibitem{bojarski2016end}
M.~Bojarski, D.~Del~Testa, D.~Dworakowski, B.~Firner, B.~Flepp, P.~Goyal, L.~D.
  Jackel, M.~Monfort, U.~Muller, J.~Zhang, \emph{et~al.}, ``End to end learning
  for self-driving cars,'' \emph{arXiv preprint arXiv:1604.07316}, 2016.

\bibitem{xu2017end}
H.~Xu, Y.~Gao, F.~Yu, and T.~Darrell, ``End-to-end learning of driving models
  from large-scale video datasets,'' in \emph{CVPR}, 2017.

\bibitem{codevilla2017end}
F.~Codevilla, M.~M{\"u}ller, A.~Dosovitskiy, A.~L{\'o}pez, and V.~Koltun,
  ``End-to-end driving via conditional imitation learning,'' \emph{arXiv
  preprint arXiv:1710.02410}, 2017.

\bibitem{codevilla2018offline}
F.~Codevilla, A.~Lopez, V.~Koltun, and A.~Dosovitskiy, ``On offline evaluation
  of vision-based driving models,'' in \emph{ECCV}, 2018.

\bibitem{tamar2016value}
A.~Tamar, Y.~Wu, G.~Thomas, S.~Levine, and P.~Abbeel, ``Value iteration
  networks,'' in \emph{NIPS}, 2016.

\bibitem{mousavian20173d}
A.~Mousavian, D.~Anguelov, J.~Flynn, and J.~Ko{\v{s}}eck{\'a}, ``3d bounding
  box estimation using deep learning and geometry,'' in \emph{CVPR}, 2017.

\bibitem{chen2016monocular}
X.~Chen, K.~Kundu, Z.~Zhang, H.~Ma, S.~Fidler, and R.~Urtasun, ``Monocular 3d
  object detection for autonomous driving,'' in \emph{CVPR}, 2016.

\bibitem{mottaghi2015coarse}
R.~Mottaghi, Y.~Xiang, and S.~Savarese, ``A coarse-to-fine model for 3d pose
  estimation and sub-category recognition,'' in \emph{CVPR}, 2015.

\bibitem{xiang2017subcategory}
Y.~Xiang, W.~Choi, Y.~Lin, and S.~Savarese, ``Subcategory-aware convolutional
  neural networks for object proposals and detection,'' in \emph{WACV}, 2017.

\bibitem{wang2018deep}
D.~Wang, C.~Devin, Q.-Z. Cai, F.~Yu, and T.~Darrell, ``Deep object centric
  policies for autonomous driving,'' in \emph{ICRA}, 2019.

\bibitem{devin2017deep}
C.~Devin, P.~Abbeel, T.~Darrell, and S.~Levine, ``Deep object-centric
  representations for generalizable robot learning,'' in \emph{ICRA}, 2017.

\bibitem{barsan2018learning}
I.~A. Barsan, S.~Wang, A.~Pokrovsky, and R.~Urtasun, ``Learning to localize
  using a lidar intensity map,'' in \emph{CORL}, 2018.

\bibitem{ushani2018feature}
A.~K. Ushani and R.~M. Eustice, ``Feature learning for scene flow estimation
  from lidar,'' in \emph{CORL}, 2018.

\bibitem{chen2017multi}
X.~Chen, H.~Ma, J.~Wan, B.~Li, and T.~Xia, ``Multi-view 3d object detection
  network for autonomous driving,'' in \emph{CVPR}, 2017.

\bibitem{luo2018fast}
W.~Luo, B.~Yang, and R.~Urtasun, ``Fast and furious: Real time end-to-end 3d
  detection, tracking and motion forecasting with a single convolutional net,''
  in \emph{CVPR}, 2018.

\bibitem{yang2018pixor}
B.~Yang, W.~Luo, and R.~Urtasun, ``Pixor: Real-time 3d object detection from
  point clouds,'' in \emph{CVPR}, 2018.

\bibitem{frossard2018end}
D.~Frossard and R.~Urtasun, ``End-to-end learning of multi-sensor 3d tracking
  by detection,'' in \emph{ICRA}, 2018.

\bibitem{dequaire2018deep}
J.~Dequaire, P.~Ondr{\'u}{\v{s}}ka, D.~Rao, D.~Wang, and I.~Posner, ``Deep
  tracking in the wild: End-to-end tracking using recurrent neural networks,''
  \emph{IJRR}, 2018.

\bibitem{yang2018hdnet}
B.~Yang, M.~Liang, and R.~Urtasun, ``Hdnet: Exploiting hd maps for 3d object
  detection,'' in \emph{CORL}, 2018.

\bibitem{liang2018deep}
M.~Liang, B.~Yang, S.~Wang, and R.~Urtasun, ``Deep continuous fusion for
  multi-sensor 3d object detection,'' in \emph{ECCV}, 2018.

\bibitem{casas2018intentnet}
S.~Casas, W.~Luo, and R.~Urtasun, ``Intentnet: Learning to predict intention
  from raw sensor data,'' in \emph{CORL}, 2018.

\bibitem{zhang2018efficient}
C.~Zhang, W.~Luo, and R.~Urtasun, ``Efficient convolutions for real-time
  semantic segmentation of 3d point clouds,'' in \emph{3DV}, 2018.

\bibitem{petrovskaya2008model}
A.~Petrovskaya and S.~Thrun, ``Model based vehicle tracking for autonomous
  driving in urban environments,'' \emph{Proceedings of Robotics: Science and
  Systems IV}, 2008.

\bibitem{darrell2001plan}
T.~Darrell, D.~Demirdjian, N.~Checka, and P.~Felzenszwalb, ``Plan-view
  trajectory estimation with dense stereo background models,'' in \emph{ICCV},
  2001.

\bibitem{palazzi2017learning}
A.~Palazzi, G.~Borghi, D.~Abati, S.~Calderara, and R.~Cucchiara, ``Learning to
  map vehicles into bird’s eye view,'' in \emph{ICIAP}, 2017.

\bibitem{pomerleau1989alvinn}
D.~A. Pomerleau, ``Alvinn: An autonomous land vehicle in a neural network,'' in
  \emph{NIPS}, 1989.

\bibitem{muller2006off}
U.~Muller, J.~Ben, E.~Cosatto, B.~Flepp, and Y.~L. Cun, ``Off-road obstacle
  avoidance through end-to-end learning,'' in \emph{NIPS}, 2006.

\bibitem{bojarski2017explaining}
M.~Bojarski, P.~Yeres, A.~Choromanska, K.~Choromanski, B.~Firner, L.~Jackel,
  and U.~Muller, ``Explaining how a deep neural network trained with end-to-end
  learning steers a car,'' \emph{arXiv preprint arXiv:1704.07911}, 2017.

\bibitem{muller2018driving}
M.~M{\"u}ller, A.~Dosovitskiy, B.~Ghanem, and V.~Koltun, ``Driving policy
  transfer via modularity and abstraction,'' \emph{arXiv preprint
  arXiv:1804.09364}, 2018.

\bibitem{chen2015deepdriving}
C.~Chen, A.~Seff, A.~Kornhauser, and J.~Xiao, ``Deepdriving: Learning
  affordance for direct perception in autonomous driving,'' in \emph{ICCV},
  2015.

\bibitem{sauer2018conditional}
A.~Sauer, N.~Savinov, and A.~Geiger, ``Conditional affordance learning for
  driving in urban environments,'' in \emph{CORL}, 2018.

\bibitem{he2017mask}
K.~He, G.~Gkioxari, P.~Doll{\'a}r, and R.~Girshick, ``Mask r-cnn,'' in
  \emph{ICCV}, 2017.

\bibitem{loper2014opendr}
M.~M. Loper and M.~J. Black, ``Opendr: An approximate differentiable
  renderer,'' in \emph{ECCV}, 2014.

\bibitem{russakovsky2015imagenet}
O.~Russakovsky, J.~Deng, H.~Su, J.~Krause, S.~Satheesh, S.~Ma, Z.~Huang,
  A.~Karpathy, A.~Khosla, M.~Bernstein, \emph{et~al.}, ``Imagenet large scale
  visual recognition challenge,'' \emph{IJCV}, 2015.

\bibitem{ross2011reduction}
S.~Ross, G.~Gordon, and D.~Bagnell, ``A reduction of imitation learning and
  structured prediction to no-regret online learning,'' in \emph{AISTATS},
  2011.

\bibitem{paszke2017automatic}
A.~Paszke, S.~Gross, S.~Chintala, G.~Chanan, E.~Yang, Z.~DeVito, Z.~Lin,
  A.~Desmaison, L.~Antiga, and A.~Lerer, ``Automatic differentiation in
  pytorch,'' in \emph{NIPS-W}, 2017.

\bibitem{krahenbuhl2018free}
P.~Kr{\"a}henb{\"u}hl, ``Free supervision from video games,'' in \emph{CVPR},
  2018.

\bibitem{lin2014microsoft}
T.-Y. Lin, M.~Maire, S.~Belongie, J.~Hays, P.~Perona, D.~Ramanan,
  P.~Doll{\'a}r, and C.~L. Zitnick, ``Microsoft coco: Common objects in
  context,'' in \emph{ECCV}, 2014.

\bibitem{Detectron2018}
R.~Girshick, I.~Radosavovic, G.~Gkioxari, P.~Doll\'{a}r, and K.~He,
  ``Detectron,'' \url{https://github.com/facebookresearch/detectron}, 2018.

\bibitem{yu2018deep}
F.~Yu, D.~Wang, E.~Shelhamer, and T.~Darrell, ``Deep layer aggregation,'' in
  \emph{CVPR}, 2018.

\bibitem{kingma2014adam}
D.~P. Kingma and J.~Ba, ``Adam: A method for stochastic optimization,''
  \emph{arXiv preprint arXiv:1412.6980}, 2014.

\bibitem{eigen2014depth}
D.~Eigen, C.~Puhrsch, and R.~Fergus, ``Depth map prediction from a single image
  using a multi-scale deep network,'' in \emph{NIPS}, 2014.

\bibitem{geiger2012we}
A.~Geiger, P.~Lenz, and R.~Urtasun, ``Are we ready for autonomous driving? the
  kitti vision benchmark suite,'' in \emph{CVPR}, 2012.

\bibitem{kim2017interpretable}
J.~Kim and J.~Canny, ``Interpretable learning for self-driving cars by
  visualizing causal attention,'' in \emph{ICCV}, 2017.

\end{thebibliography}
}
\end{spacing}
\end{document}